\documentclass[preprint,12pt]{elsarticle}

\usepackage{amssymb}
\usepackage{amsmath}
\usepackage{subfig}
\usepackage{bm}
\usepackage{multirow}

\usepackage{graphicx}
\usepackage{booktabs}
\usepackage{diagbox}
\usepackage{xcolor}
\usepackage{lineno,hyperref}
\usepackage{url}

\journal{Neural Networks}

\begin{document}

\begin{frontmatter}



\title{Achieving Rotation-Invariant Convolution via Non-Learnable Orientation Alignment Operators}


\author{Hanlin Mo$^{a}$, Peihong Lei$^{b}$, You Hao$^{c}$, Guoying Zhao$^{d}$\corref{cor1}} 

\cortext[cor1]{The corresponding author}
\affiliation{organization={Unmanned System Research Institute, Northwestern Polytechnical University},
	state={Xi'an},
	postcode={710072}, 
	country={China}}
\affiliation{organization={School of Artificial Intelligence, Xidian University},
	        state={Xi'an},
            postcode={710126}, 
            country={China}}
\affiliation{organization={Key Laboratory of Intelligent Information Processing, Institute of Computing Technology, Chinese Academy of Sciences},
	state={Beijing},
	postcode={100190}, 
	country={China}}
\affiliation{organization={Center of Machine Vision and Signal Analysis, University of Oulu},
	state={Oulu},
	postcode={90570}, 
	country={Finland}}
\vspace{10mm}
\begin{abstract}
Achieving rotational invariance in deep neural networks without data augmentation is a research hotspot. Intrinsic invariance enables features to capture targets' inherent properties, enhancing deep learning performance in visual tasks. Based on various types of non-learnable operators, this paper proposes a comprehensive set of convolution operations that are natually invariant to arbitrary rotations. Unlike most prior methods, these rotation-invariant convolutions (RIConvs) have the same number of learnable parameters and a similar computational process as standard convolutions, making them interchangeable. Using the MNIST-Rot dataset, we validate their invariance across rotation angles and compare them with previous rotation-invariant CNNs, where two gradient-based RIConvs achieve state-of-the-art results. Then, we integrate RIConvs with classic CNN backbones and evaluate them on texture recognition, aircraft type recognition, and remote sensing image classification tasks. Results show that RIConvs significantly improve accuracy, particularly with limited training data, and enhance performance even with data augmentation. Our codes can be downloaded from \url{https://github.com/HanlinMo/RIConvs.git}.  
\end{abstract}




\end{frontmatter}


\section{Introduction}
\label{Section1}
In simple terms, the core challenge in the field of computer vision is how to enable machines to process visual information they acquire efficiently, much like the human visual system. Since 2012, Deep Neural Networks (DNNs), represented by Convolutional Neural Networks (CNNs), have made remarkable advancements in core tasks of computer vision and pattern recognition, such as object classification \cite{1,2}, retrieval\cite{3,4}, detection \cite{5,6}, and segmentation \cite{7,8}. This has led people to question whether the core challenge mentioned above has already been addressed. In reality, there is still a huge gap between current DNN models and the human brain, and we choose an important property, rotation invariance, to illustrate this issue. 

Rotation invariance is one of the most common type of perceptual constancy, which enables the human visual system to efficiently and accurately recognize objects with arbitrary orientations, and plays a vital role in numerous practical applications, such as textural image analysis \cite{9}, remote sensing image classification \cite{10,11}. However, CNN models lack the invariance to image rotations, and simply increasing the number of model layers or learnable parameters does not effectively enhance the rotation invariance of CNNs. A common solution is to use data augmentation techniques to rotate input samples during DNN training, resulting in data-driven rotation-invariant CNNs (RI-CNNs). However, previous research has indicated that data augmentation further weakens the interpretability of CNNs that are already "black boxes", disrupts semantic information of data, and reduces the effective utilization of learnable parameters, leading to increased redundancy \cite{12,13}. Clearly, it is not an ideal solution for achieving rotation invariance in CNNs. 

Thus, recent research has focused on ensuring the rotation invariance of CNNs through the mechanisms within the models themselves rather than relying on data. Representative works include the Spatial Transformer Network (STN) \cite{14}, Polar Transformer Network \cite{15}, Group Equivariant Convolutional Network (G-CNN) \cite{16}, Oriented Response Network (ORN) \cite{17}, Harmonic Network (H-Net) \cite{13}, and General E(2)-Equivariant Steerable CNN (E(2)-CNN) \cite{18}, among others \cite{19,20,21,22}. However, they suffer from the following three issues: \textbf{(1)} Most of them exhibit invariance only to specific rotation angles and cannot handle continuous/arbitrary rotation angles \cite{16,17,19}. \textbf{(2)} Some rotation-invariant network modules have architectures that are completely different from conventional convolution operations, making them non-interchangeable \cite{12,20,23}. This prevents them from being integrated with classic CNN backbones such as VGG \cite{24} and ResNet \cite{25}. \textbf{(3)} Due to the inclusion of additional learnable parameters, training of some methods still relies on data augmentation \cite{10,14,26}.

The most ideal scenario would be to design new convolution operations that can be interchanged with conventional convolution operation without increasing the number of learnable parameters and naturally exhibits invariance to arbitrary rotations. In fact, researchers have begun exploring this direction. Recently, we designed three rotation-invariant convolution operations (RIConvs) based on Sobel operator\cite{27}, rotation-invariant coordinate systems\cite{28}, and sort operation\cite{29}, meeting the aforementioned requirements. Also, Hao et~al. proposed another similar RIConv using Local Binary Pattern (LBP) descriptor \cite{30}. 

However, these studies are still preliminary and not systematic. Firstly, many other methods can also be used to achieve rotation invariance in convolution operations, such as maximum operation and gradient estimation based on first-order Gaussian derivatives. Secondly, the performance of RIConvs implemented in different ways has not been compared in practical tasks. Previous works have also not analyzed the advantages and inherent limitations of various RIConv. Finally, we need to address whether the natural rotational invariance of convolutions can further enhance the performance of CNN models when data augmentation is used. The purpose of this paper is to address these issues. Our main contributions are as follows:
\begin{itemize}
\item We design seven RIConvs using Sobel operator, Gaussian derivatives, maximum operator, LBP descriptor and other non-learnable operators. Four of them are proposed for the first time in this paper, while the other two are improvements based on our previous work. Each of RIConvs has the same number of learnable parameters as its corresponding conventional convolution operation, and the output feature maps of both also have the same size, thus enabling them to be interchangeable.
\item Without using data augmentation, we validate the invariance of different RIConvs to various rotation angles using the MNIST-rot dataset \cite{31}, and compare their performance with previous rotation-invariant CNN models. We found that the performance of some RIConvs was unstable and identified their inherent limitations. 
\item  We combine these RIConvs with classical CNN backbones of different types and depths (such as VGG\cite{24}, Inception\cite{32}, ResNet\cite{25}, and DensNet\cite{33}), and test their performance in texture image classification, aircraft type recognition, and remote sensing image classification based on Outex\_TC\_00012\cite{34}, MTARSI-20\cite{35}, and NWPU-RESISC45 datasets\cite{36}. Also, we analysis the influence of data augmentation on RIConvs.  
\end{itemize}

\section{Related Work}
\label{Section2} 
This section briefly introduces the existing methods for achieving rotation invariance in CNN models, most of which can be categorized into the following three types. 

\subsection{Data-driven RI-CNNs}
\label{Section2.1}
Using data augmentation is the most direct method to achieve rotation invariance in CNN models, namely applying random rotations to input images during model training. However, previous research has found that using data augmentation can lead to a significant amount of learnable parameter redundancy in the model \cite{13}. As the model learns from the training data that it should extract multi-orientation information, there inevitably are different rotation "copies" of the same convolutional kernel in its convolutional layers. Clearly, these redundant parameter "copies" are the cost the model incurs to achieve rotation invariance. Additionally, some researchers have found that data augmentation further diminishes the feature interpretability of CNNs\cite{12}. 

Beyond direct data augmentation, some methods use "implicit" data augmentation to achieve rotation-invariant networks \cite{14,22,24}. For instance, TI-Pooling \cite{24} first generates multiple rotated versions of the input image, then uses the same CNN to extract their features separately, and performs a pooling operation on these features. Methods like STN \cite{14,37} essentially belong to this category too. These methods first uses learnable modules to estimate rotation angle of the input, and then calibrate the input to a standard orientation using the angle before extracting features. However, to enable the angle estimation module to predict correct rotation angles, we need to train it using samples which are randomly rotated. In addition, such methods generally can only handle global rotations of images, not local rotations (i.e., only some objects rotate while other obejcts and background remain unchanged). Moreover, some utilize Siamese structure and contrastive learning loss to achieve the model's rotation invariance \cite{21}, but also employ data augmentation when constructing positive sample pairs.

\subsection{Rotation-equivariant CNNs}
\label{Section2.2}
Based on the theories of Lie groups and Lie algebras, some researchers have designed rotation-equivariant networks represented by group convolution networks\cite{16} and E(2)-CNN\cite{18}, and have successfully applied them to medical image analysis and protein structure prediction\cite{9,38}. Taking group convolution as an example, it extracts features of the input in multiple orientations simultaneously. This set of multi-orientation features forms the group features of the input, which exhibits equivariance rather than invariance to rotations, and is input as a whole into the next group convolution layer. However, group convolution generally only has invariance to specific rotation angles, such as multiples of $90^{\circ}$. It should be noted that the structure of equivariant convolutions often differs significantly from standard convolutions, making their integration into classic CNN backbones quite challenging. Moreover, the computational cost of some such methods is relatively high\cite{27}.

\subsection{Mechanism-assured RI-CNNs}
\label{Section2.3}
Due to various issues with data-driven RI-CNNs, some researchers have started exploring certain mechanisms to ensure CNNs' rotation invariance, proposing different solutions. For example, some researchers use the representation of inputs in polar or log-polar coordinates as the input for CNNs\cite{15,39}. In fact, 2D rotations in Cartesian coordinates become translations in log-polar/polar coordinates, and traditional convolution operations are invariant to 2D translations. However, the transformation of coordinates disrupts the spatial relationships between image contents, leading to a decline in model performance. Moreover, similar to STNs\cite{14}, these methods can only handle global rotations of images. ORN\cite{17} and Rotation Equivariant Vector Field Network (RotEqNet)\cite{19} extract features of the input in multiple orientations through rotating convolutional kernels. For example, if convolitional kernels are rotated by $k\cdot45^{\circ}, k=1, 2, ..., 8$, these models are invariant to those specific rotation angles. Clearly, to achieve rotation invariance at any angle, we need to generate as many rotated copies of the convolutional kernel as possible, which significantly increases the model's computational efficiency. 

Recent work has begun to explore how to achieve invariance of CNN models for continuous rotation angle, and requires that the designed RIConvs and conventional convolution can be replaced with each other. To this end, we designed Gradient-Align CNN using Sobel operator\cite{25}, rotation-invariant coordinate CNN\cite{27} using rotation-invariant coordinates systerm, and sorting convolution \cite{28} using ring sorting operations. Hao et al. designed the Regional Rotate Layer using LBP\cite{26}. In fact, these methods share a similar computational process, which involves calibrating the local region involved in the convolution operation with a non-learnable operator first, followed by the standard convolution operation. Since non-learnable operators are used, these RIConvs have the same number of learnable parameters as their corresponding traditional convolutions. Furthermore, their computational processes are similar, allowing them to be interchangeable. These methods form the research foundation and starting point of our paper. In this paper, we design a more complete set of new RIConvs using various types of non-learnable operators. Based on the experimental results obtained on synthetic and real-world image datasets, we systematically compare the performance of different RIConvs on various tasks and identified some inherent shortcomings in certain RIConvs.

\section{Methodology}
\label{Section3}
This section first illustrates how to achieve the invariance of conventional convolution operation under any rotation angles using gradient operators, sorting operation, LBP feature, and maximum operation, respectively. Then, we explain how to combine the designed RIConvs with classical CNN backbones.

\subsection{Implementing RIConv Using Non-Learnable Operations}
\label{Section3.1}
The input of a convolution operation is typically an image or a feature map, which can be represented as a $h\times w \times c$ tensor, where $h$, $w$, and $c$ represent the height, width, and number of channels of the tensor, respectively. To simplify subsequent analysis, we assume $c=1$. In this way, the input tensor can be represented as a 2D function $F(X):\Omega \rightarrow\mathbb{R}$, where the domain $\Omega = {1,2,...,h}\times{1,2,...,w}$. At a position $X_{0}\in\Omega$, the conventional convolution operation \emph{Conv} can be expressed as
\begin{equation}\label{equ:1}
	Conv(X_{0},F(X))=\sum_{P\in\mathcal{S}}W(P)\cdot F(X_{0}+P)
\end{equation}
where $W$ is a $K\times K$ learnable kernel, $K$ is a non-negative integer; $P$ enumerates all sampling positions on a $K\times K$ square grid $\mathcal{S}$ where $X_{0}$ is the origin center of the grid. For example, when $W$ is a $3 \times 3$ kernel, we have $\mathcal{S}={(-1,-1),(-1,0), \cdots, (0,1),(1,1)}$, which contains $9$ sample positions. 

Let $G(Y)$ be a rotated version of $F(X)$, i.e., $G(Y) = F(R_{-\theta}Y)$, where $R_{-\theta}$ is a $2\times2$ rotation matrix and $\theta\in[0, 2\pi)$ represents the rotation angle. Assuming $Y_{0}$ is the corresponding position of $X_{0}$ after rotation, the convolution operation at $Y_{0}$ is
\begin{equation}\label{equ:2}
	Conv(Y_{0},G(Y))=\sum_{P\in\mathcal{S}}W(P)\cdot G(Y_{0}+P)
\end{equation}
Since $G(Y)=F(R_{-\theta}Y)$, we have
\begin{equation}\label{equ:3}
	G(Y_{0}+P)=F(R_{-\theta}(Y_{0}+P))=F(X_{0}+R_{-\theta}P)
\end{equation}
By substituting (\ref{equ:3}) into (\ref{equ:2}), we can find that 
\begin{equation}\label{equ:4}
	\begin{split}
		Conv(Y_{0},G(Y))&=\sum_{P\in\mathcal{S}}W(P)\cdot F(X_{0}+R_{-\theta}P)\\&\ne \sum_{P\in\mathcal{S}}W(P)\cdot F(X_{0}+P) \ne Conv(X_{0},F(X))
	\end{split}
\end{equation}
This means that Conv is not invariant to 2D rotations.

In fact, a direct and simple way to achieve the rotation invariance of Conv is to eliminate the influence of rotations on the $K\times K$ region involved in the convolution operation. In this paper, we achieve this through various non-learnable operations (NLOPs). For example, we can utilize certain gradient operators to compute the gradient direction of the region, and eliminate the influence of the rotation angle $\theta$ by aligning the region to a standard orientation. Therefore, the proposed RIConv can be defined as:
\begin{equation}\label{equ:5}
	RIConv(X_{0},F(X))=\sum_{P\in\mathcal{S}}W(P)\cdot NLOP(F(X_{0}+P))
\end{equation}
For any $P\in\mathcal{S}$, $NLOP$ ensures the following relationship always holds: 
\begin{equation}\label{equ:6}
NLOP(G(Y{0}+P))=NLOP(F(X_{0}+P))
\end{equation}
thus ensuring $RIConv(G(Y),Y_{0})=RIConv(F(X),X_{0})$.

We take a $3\times3$ convolution as an example to introduce seven types of NLOPs that can eliminate the influence of rotations. It is worth noting that compared to the Cartesian coordinate system, our previous studies have shown that RIConvs implemented based on the polar coordinate system have better invariance\cite{27,28}. Therefore, we adopt this approach as well. For a $3\times3$ region under the polar coordinate system, apart from $X_{0}$, we need to sample 8 positions, $P_{1}, P_{2}, \cdots, P_{8}$, at equal angular intervals on the circumference with $X_{0}$ as the center and a radius of $1$. The angle corresponding to the position $P_{i}$ is $(i-1)\cdot\frac{360^{\circ}}{8}$ (clockwise direction), where $i=1,2,\cdots,8$. Using bilinear interpolation, we can obtain the pixel values of the input $F(X)$ at these positions, which will participate in the convolution operation.

\begin{figure}
	\centering
	\subfloat[SB-Conv]
	{\includegraphics[height=28mm,width=100mm]{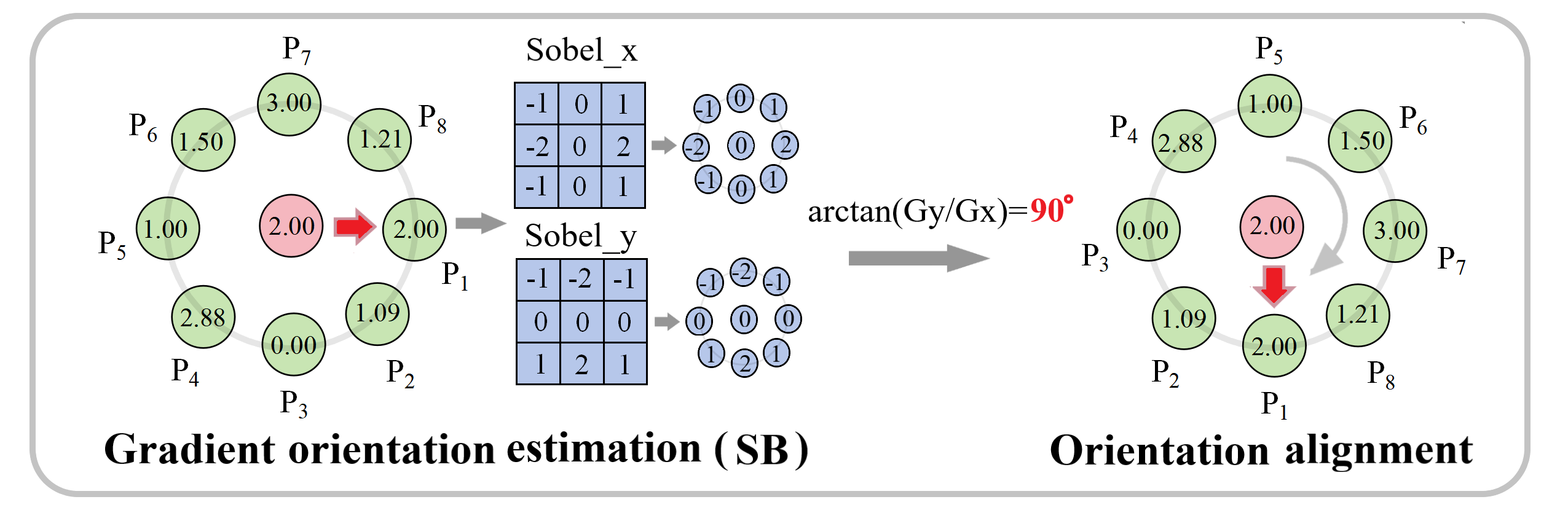}\label{figure:1(a)}}\\
	\subfloat[GD-Conv]
	{\includegraphics[height=28mm,width=100mm]{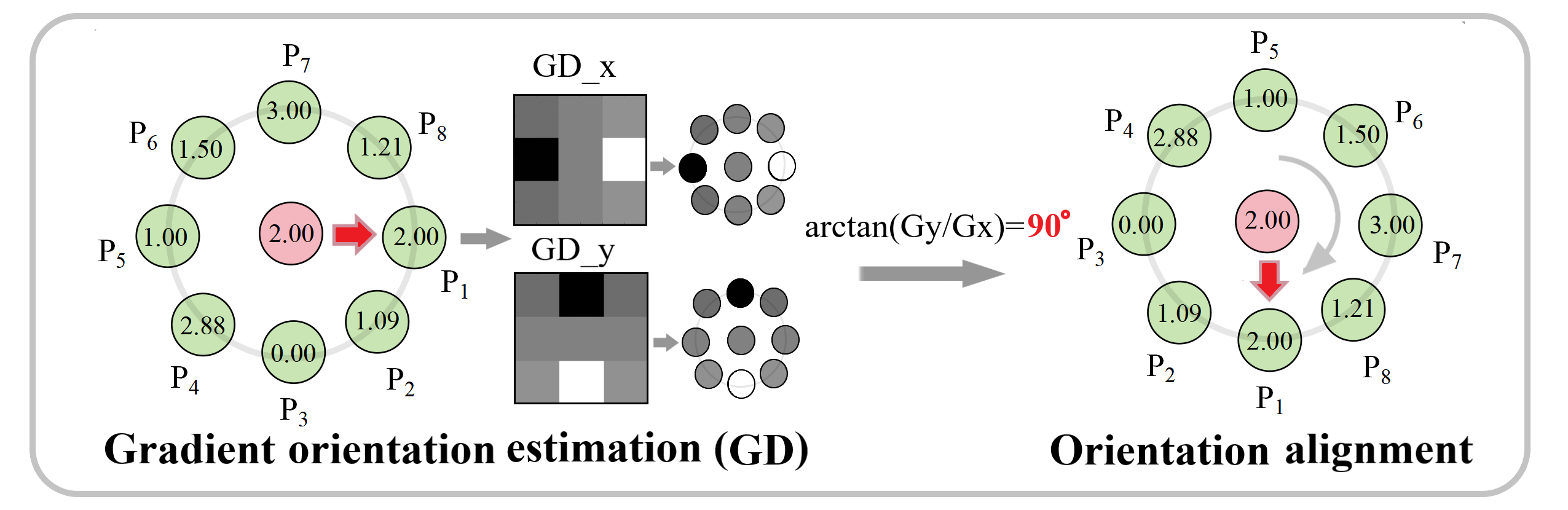}\label{figure:1(b)}}\\
	\subfloat[ST-Conv]
	{\includegraphics[height=28mm,width=100mm]{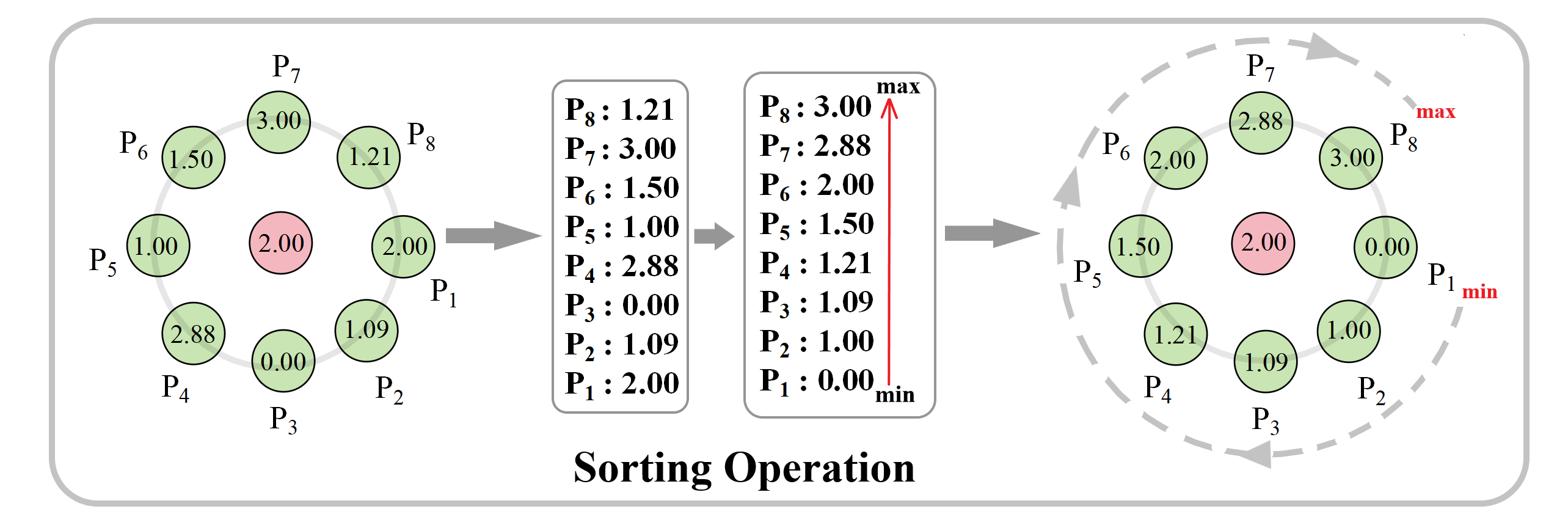}\label{figure:1(c)}}\\
	\caption{Illustration of how to use the Sobel operator, the first-order Gaussian derivative, and the sorting operator to eliminate the effects of arbitrary rotations acting on a $3\times3$ neighborhood. The RIConvs implemented using these tree NLOPs are SB-Conv, GD-Conv, and ST-Conv, respectively.}\label{figure:1}
\end{figure}

\textbf{SB-Conv:} SB-Conv uses $3\times3$ Sobel operators to compute the gradient of a given region, and then rotates the region to a standard orientation using the obtained gradient orientation. Specifically, $3\times3$ Sobel operators on $x$ and $y$ directions are defined as follows:
\begin{equation}\label{equ:7}
	Sobel\_x=\left(
	\begin{array}{ccc}
		-1& 0& 1\\
		-2& 0& 2\\
		-1& 0& 1
	\end{array}
	\right),
	Sobel\_y=\left(
	\begin{array}{ccc}
		-1& -2& -1\\
		0& 0& 0\\
		1& 2& 1
	\end{array}
	\right)
\end{equation}
We use a simple nearest-neighbor interpolation instead of bilinear interpolation to transform $Sobel\_x$ and $Sobel\_y$ into the polar coordinate system. Then, we convolve them separately with the $3\times3$ region to obtain the gradient $Gradient=[G_{x}, G_{y}]$ of that region. Subsequently, the gradient orientation $\phi\in(0,\pi]$ is calculated using the arctan function $\phi=arctan(G_{y}/G_{x})$, and the region is rotated to the standard orientation (i.e., the gradient orientation coincides with the horizontal axis direction) using this angle. Fig.\ref{figure:1(a)} illustrates this process more clearly. When the region is rotated by any angle $\theta$, the gradient orientation calculated using Sobel operators also becomes $(\phi+\theta)$. Therefore, the region is always aligned to the same standard orientation. Since the convolution operation is performed on the aligned region, SB-Conv evidently exhibits rotation invariance.  

\textbf{GD-Conv:} 
Similar to SB-Conv, GD-Conv also calibrates the $3\times3$ region using gradient orientation to achieve rotation invariance of convolution. The only difference is that we use the first derivatives of 2D Gaussian function $GD\_x$ and $GD\_y$ instead of the Sobel operator to compute the gradient (see Fig.\ref{figure:1(b)}). The definitions of $GD\_x$ and $GD\_y$ are as follows:
\begin{equation}\label{equ:8}
	GD\_x=\frac{-1}{2\pi\sigma^{4}}\cdot x\cdot e^{-\frac{x^{2}+y^{2}}{2\sigma^{2}}},
	GD\_y=\frac{-1}{2\pi\sigma^{4}}\cdot y\cdot e^{-\frac{x^{2}+y^{2}}{2\sigma^{2}}}
\end{equation}
where the parameter $\sigma$ is the scale factor used to control the decay rate of the Gaussian function. To ensure that all information of $GD\_x$ and $GD\_y$ falls within the $K\times K$ region as much as possible, according to the three-sigma rule, we set $\sigma=K/6$. Therefore, when the region size is $3\times3$, $\sigma=0.5$. Additionally, due to the smoothing effect of the Gaussian function, the first-order Gaussian derivative is more robust to disturbances such as noise compared to Sobel operators. 

\textbf{ST-Conv:} ST-Conv uses a sorting operation to eliminate the influence of rotation. As shown in Fig.\ref{figure:1(c)}, we first sort pixel values of the input at the other 8 sampling positions $[F(X_{0}+P_{1}),...,F(X_{0}+P_{8})]$ in ascending order, except for the central position $X_{0}$. Then, the sorted values are placed back at positions $P_{1}, P_{2}, ..., P_{8}$ in order, with the minimum value placed at $P_{1}$ and the maximum value at $P_{8}$. Obviously, even if the local region undergoes rotation, the sorted sequence of values remains unchanged, ensuring the rotation invariance of ST-Conv. It should be noted that the sorting operation actually disrupts the local structure of the input to some extent, leading to information loss, especially when dealing with regions of large size $K$. To solve this problem, we proposed the ring sorting operation in \cite{28}. In simple terms, when implementing convolution kernels of size $5\times 5$ and $7\times 7$, we sort the sampling positions with different radial values separately, to preserve some spatial information as much as possible.

\begin{figure}
	\centering
	\subfloat[LBP-Conv]
	{\includegraphics[height=27mm,width=133mm]{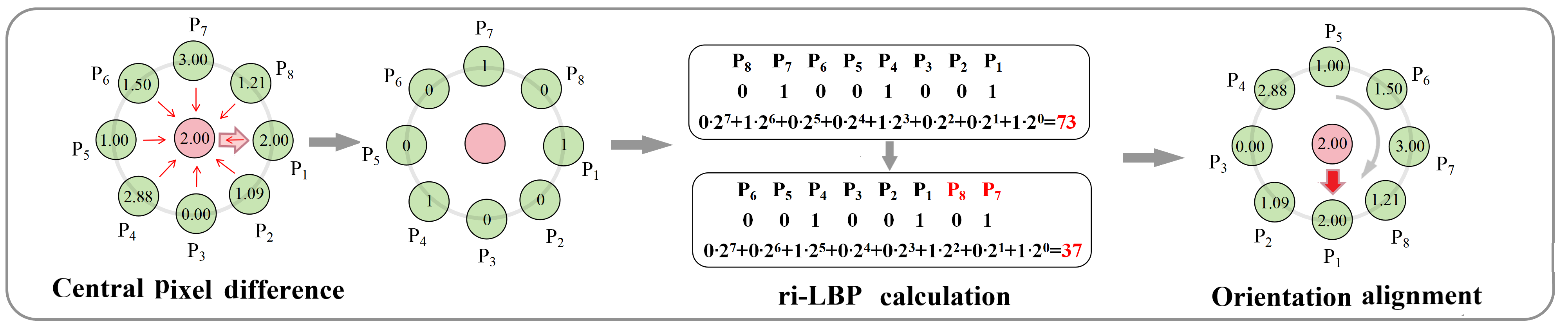}\label{figure:2(a)}\hfill}\\
	\subfloat[$\overline{\mathrm{LBP}}$-Conv]
	{\includegraphics[height=27mm,width=133mm]{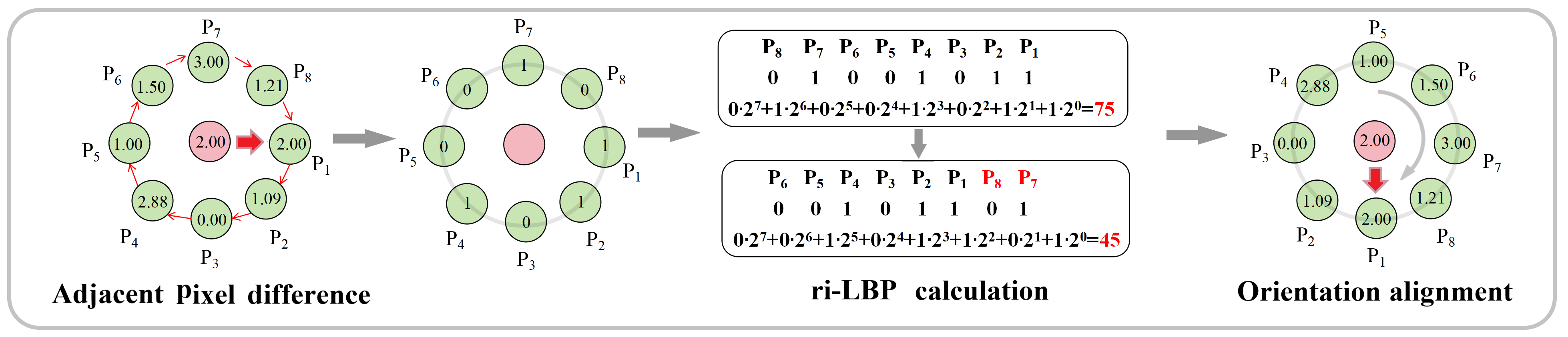}\label{figure:2(b)}\hfill}
	\caption{Illustration of how to use the LBP operator to eliminate the effects of arbitrary rotations acting on a 3x3 neighborhood. Based on central pixel difference and adjacent pixel difference, we define two different types of LBP, which are used to implement LBP-Conv and $\overline{\mathrm{LBP}}$-Conv, respectively. }\label{figure:2}
\end{figure}

\textbf{LBP-Conv:} This method first computes the pixel differences between the input at the sampling position $P_{1}\sim P_{8} $ and the center position $X_{0} $, i.e., $[F(X_{0}+P_{1})-F(X_{0}), F(X_{0}+P_{2})-F(X_{0}),\cdots, F(X_{0}+P_{8})-F(X_{0})]$ (see Fig.\ref{figure:2(a)}). Then, the pixel differences are binarized, where values less than $0$ are set to $0$, and values greater than or equal to $0$ are set to $1$, resulting in a binary encoding called LBP feature, for example, $01001001$ (from $P_{8}$ to $P_{1}$). To achieve LBP feature's rotation invariance, the leading bit of the binary descriptor needs to be continuously moved to the end, and the corresponding decimal value of the new binary encoding is computed. By repeating this process, the smallest decimal value and its corresponding binary encoding can be found. For example, the decimal value corresponding to $01001001$ is $73$, and after moving its leading bit to the end twice, we get $00100101$, which corresponds to the smallest decimal value $37$. At this point, the sampling positions corresponding to this new binary number are $P_{6},P_{5},P_{4},P_{3},P_{2},P_{1},P_{8},P_{7}$. That is, $P_{7}$ is rotated to the original position of $P_{1}$. Clearly, through this method, we can also rotate the region to the standard orientation, eliminating the influence of rotations.

\bm{$\overline{\bm{\mathrm{LBP}}}$}\textbf{-Conv:} We design a variant of LBP-Conv, whose calculation process is basically the same as LBP-Conv, except that we do not calculate the pixel difference between other positions and the center position, but instead calculate the pixel difference between adjacent positions, i.e., $[F(X_{0}+P_{1})-F(X_{0}+P_{2}), F(X_{0}+P_{2})-F(X_{0}+P_{3}),\cdots, F(X_{0}+P_{8})-F(X_{0}+P_{1})]$ (see Fig.\ref{figure:2(b)}). In fact, previous researchers have designed various variants of standard LBP using different encoding schemes, and we plan to compare the invariance of RIConvs implemented using different LBP encodings.

\begin{figure}
	\centering
	\subfloat[MAX-Conv]
	{\includegraphics[height=29mm,width=59mm]{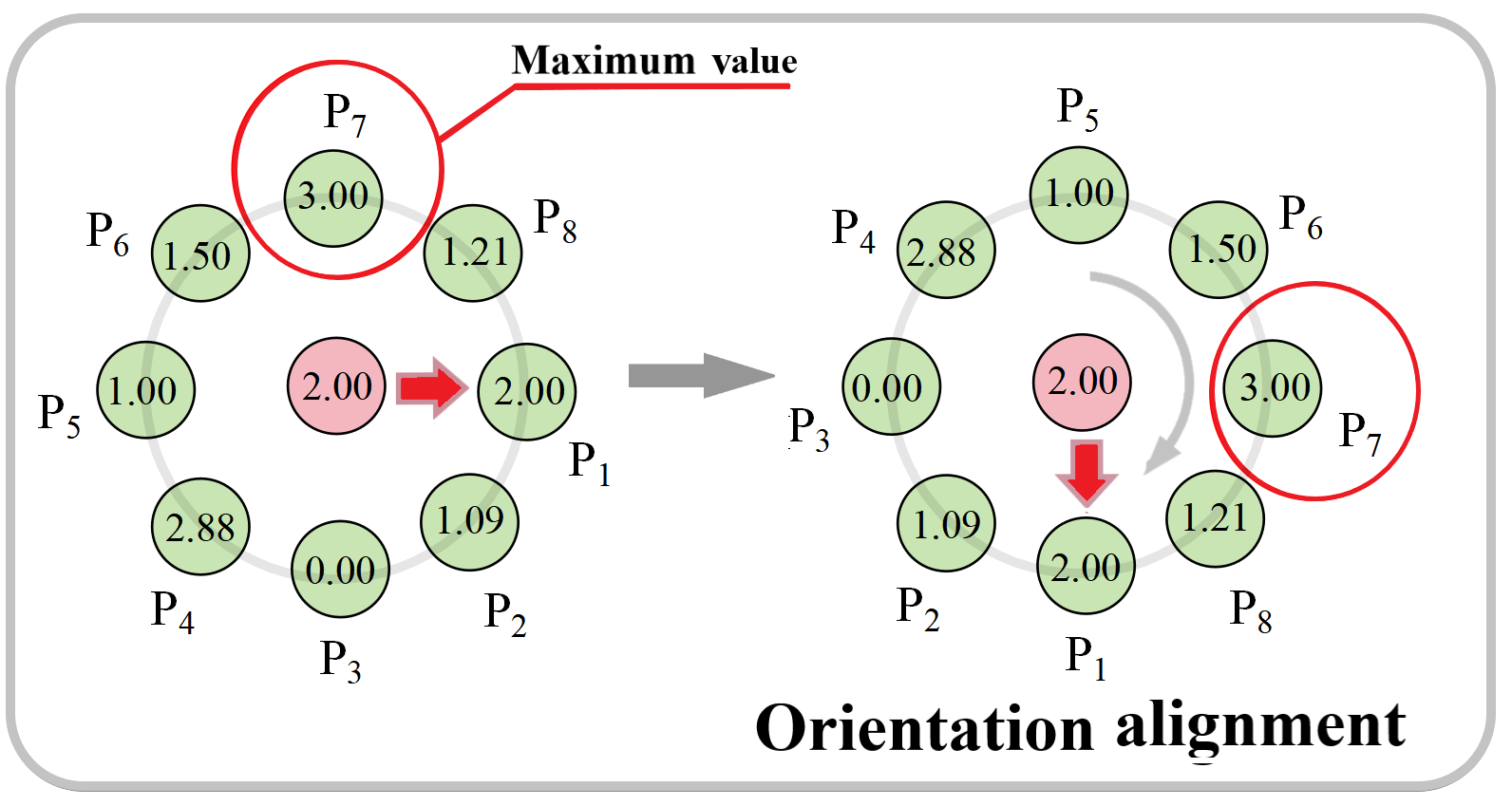}\label{figure:3(a)}\hfill~}
	\subfloat[$\overline{\mathrm{MAX}}$-Conv]
	{\includegraphics[height=29mm,width=79mm]{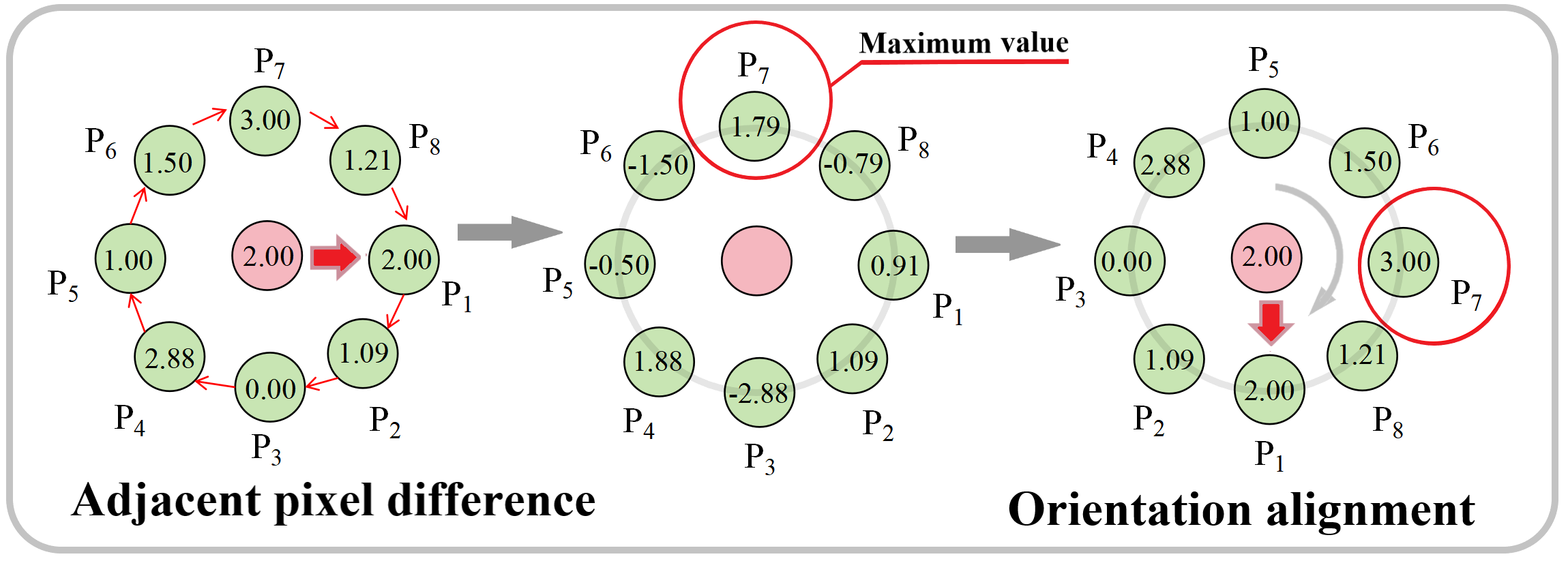}\label{figure:3(b)}}
	\caption{Illustration of how to use the maximum operator to eliminate the effects of arbitrary rotations acting on a $3\times3$ neighborhood. By aligning the maximum value of pixels and adjacent pixel differences to a standard position, we design two RIConvs, MAX-Conv and $\overline{\mathrm{MAX}}$-Conv, respectively.}\label{figure:3}
\end{figure}

\textbf{MAX-Conv:} For the pixel values at the 8 sampling positions around the center point, $[F(X_{0}+P_{1}),...,F(X_{0}+P_{8})]$, MAX-Conv first identifies the position of the maximum value, for example, $P_{7}$. Suppose a $3\times3$ region involved in the convolution computation is rotated by $90^{\circ}$, the maximum value within the region remains unchanged, but its position changes from $P_{7}$ to $P_{9}$ (rotating $P_{7}$ clockwise by $90^{\circ}$ will aligns it with $P_{9}$). Clearly, by rotating the position where the maximum value occurs to the standard direction (aligning it with the horizontal direction), we can achieve the convolution's rotation invariance. The entire process is illustrated in Fig.\ref{figure:3(a)}. 

\bm{$\overline{\bm{\mathrm{MAX}}}$}\textbf{-Conv:} Similar to $\overline{\mathrm{LBP}}$-Conv, $\overline{\mathrm{MAX}}$-Conv first computes the pixel differences at adjacent sampling positions of the input, i.e., $[F(X_{0}+P_{1})-F(X_{0}+P_{2}), F(X_{0}+P_{2})-F(X_{0}+P_{3}),\cdots, F(X_{0}+P_{8})-F(X_{0}+P_{1})]$; then identifies the position where the maximum difference occurs and rotates that position to the standard orientation (see Fig.\ref{figure:3(b)}). In fact, MAX-Conv can only be used if there is only one maximum value within the $K\times K$ region. If there are multiple identical maximum values within the region, the positions where two $K\times K$ regions satisfying the rotation relationship are aligned to the standard orientation may differ. This weakens the rotation invariance of MAX-Conv. The introduction of $\overline{\mathrm{MAX}}$-Conv is to mitigate this problem to some extent, as the probability of multiple identical pixel differences occurring is relatively low. 

To summarize, among seven types of RIConvs designed above, GD-Conv, $\overline{\mathrm{LBP}}$-Conv, Max-Conv and $\overline{\mathrm{MAX}}$-Conv are proposed for the first time in this paper. We have proposed SB-Conv and ST-Conv in \cite{27,29}. However, the original SB-Conv was calculated based on the Cartesian coordinate system, while in this paper, we redefine it using the polar coordinate system. Our experimental results will demonstrate that this modification significantly improves its performance. LBP-Conv was proposed by Hao et al. \cite{30}. Similar to SB-Conv, we also calculate it using the polar coordinate system in this paper. Furthermore, in Section \ref{Section4.2}, we analyze the intrinsic shortcomings of LBP-Conv, explain why $\overline{\mathrm{LBP}}$-Conv can partially address these issues, and provide some experimental results to substantiate this point.  

\subsection{Integrating RIConv with CNNs}
\label{Section3.2}
Obviously, all RIConv proposed in the previous section have the same number of learnable parameters as their corresponding Conv. For an input, if the outputs of these RIConvs are of the same size as the output of the Conv, it indicates that they can be mutually substituted. Then, we can further achieve a CNN backbones's rotation invariance by replacing its all Convs with a certain type of RIConvs. Assuming the height and width of an input are $h$, $w$, and the kernel size of both RIConvs and the Conv is $K\times K$. When using zero padding during computation, the output size of the Conv is $h\times w$. For RIConvs, before performing the convolution operation, we first need to align $K\times K$ local region at each spatial position $X\in\Omega$ using non-learnable operators. Then, we concatenate all calibrated local regions according to their spatial positions to form a new input, where its height and width are $(K\cdot h)$ and $(K\cdot w)$, respectively. Next, we perform $K\times K$ convolution operation on this input but set the convolution stride to $K$ and do not use padding. Obviously, the output size obtained with this setting is also $h\times w$, which is exactly the same as that obtained with the Conv. Therefore, RIConvs and Conv can replace each other.

\begin{figure}
	\centering
	\subfloat[The calculation process of $3\times3$ Conv on a $5\times5$ input.]
	{\includegraphics[height=28mm,width=120mm]{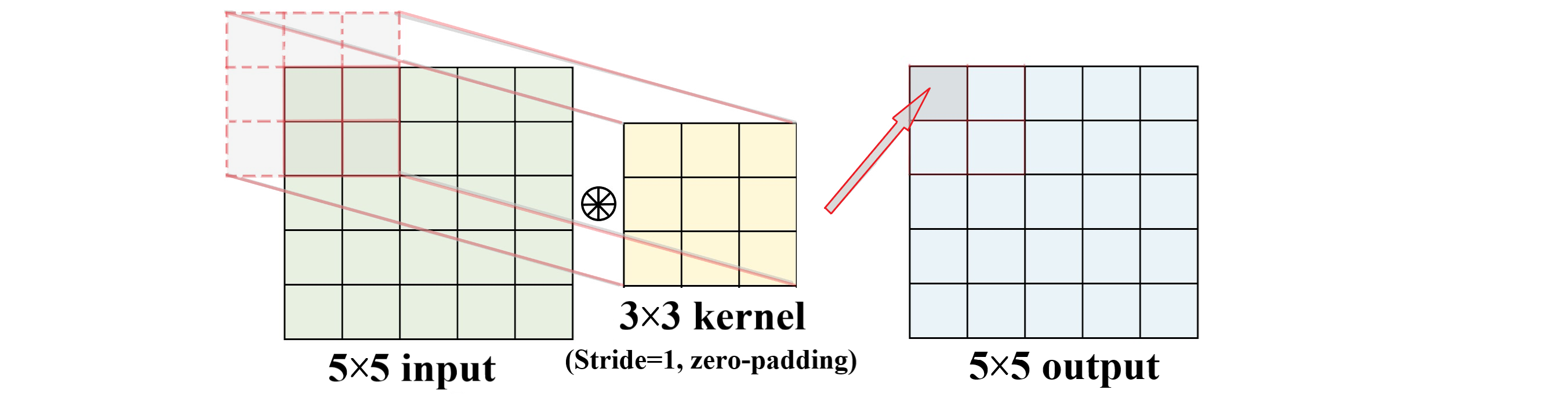}\label{figure:4(a)}}\\
	\subfloat[The calculation process of $3\times3$ RIConv on a $5\times5$ input.]
	{\includegraphics[height=28mm,width=130mm]{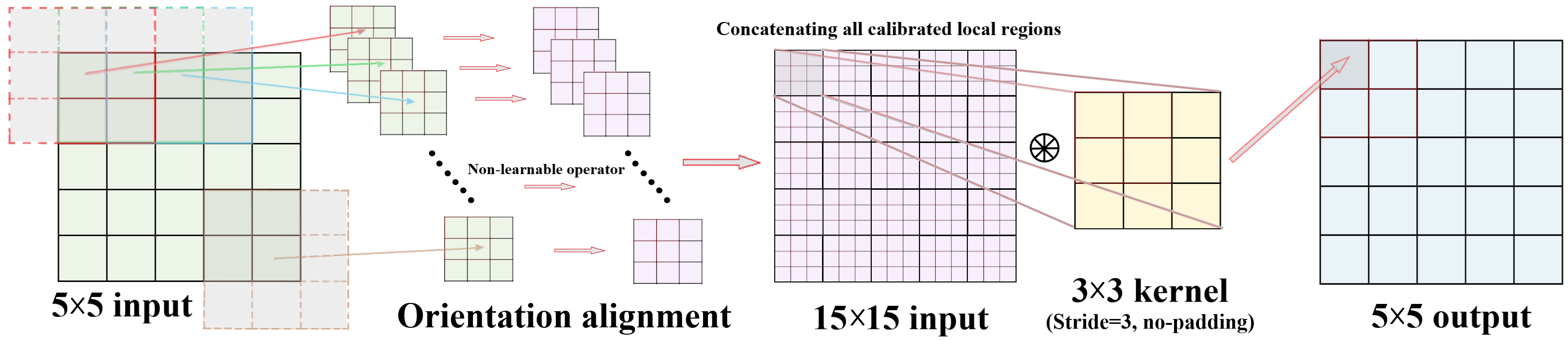}\label{figure:4(b)}}
	\caption{Using a $3\times3$ convolutional kernel and a $5\times5$ input as an example, this figure illustrates that when the input is identical, the outputs obtained using RIConvs and their corresponding Convs have the same size.}\label{figure:4}
\end{figure}

By replacing all Conv in a traditional CNN models with some type of RIConvs, we can obtain a RI-CNN. When simultaneously inputting an image and its rotated version into the RI-CNN, the two features obtained in each RIConv layer satisfy the same rotation relationship. If we use max or average pooling operations to downsample the spatial size of this output to $1\times1$ (downsampling can also be achieved in previous RIConv layers by setting convolution stride $>1$ or using pooling operations), the resulting feature is invariant to any rotations and can be further used as the input of fully connected layers.  

\section{Experiments}
\label{Section4}
This section validates the performance of RIConvs in various practical tasks. Based on the MNIST-Rot dataset, we first train various types of RIConvs without using data augmentation, and then test their performance on test sets that contains rotated images to evaluate their rotation invariance, and compare their performance with some existing RI-CNNs. We identify the inherent limitations of some RIConvs and explained the reasons behind their unstable performance. Subsequently, we deploy RIConvs into classical CNN backbones and test their performance on texture image recognition, aircraft classification, and remote sensing image tasks based on the Outex\_00012, MTARSI, and RESISC-45 datasets. Finally, we demonstrate through experiments that RConvs can still further improve the performance of CNN models even when using data augmentation. 

\subsection{Experiment Setup}
\label{Section4.1}
\begin{figure*}
	\centering
	\subfloat[The MNIST-Rot dataset.]
	{\includegraphics[height=45mm,width=64mm]{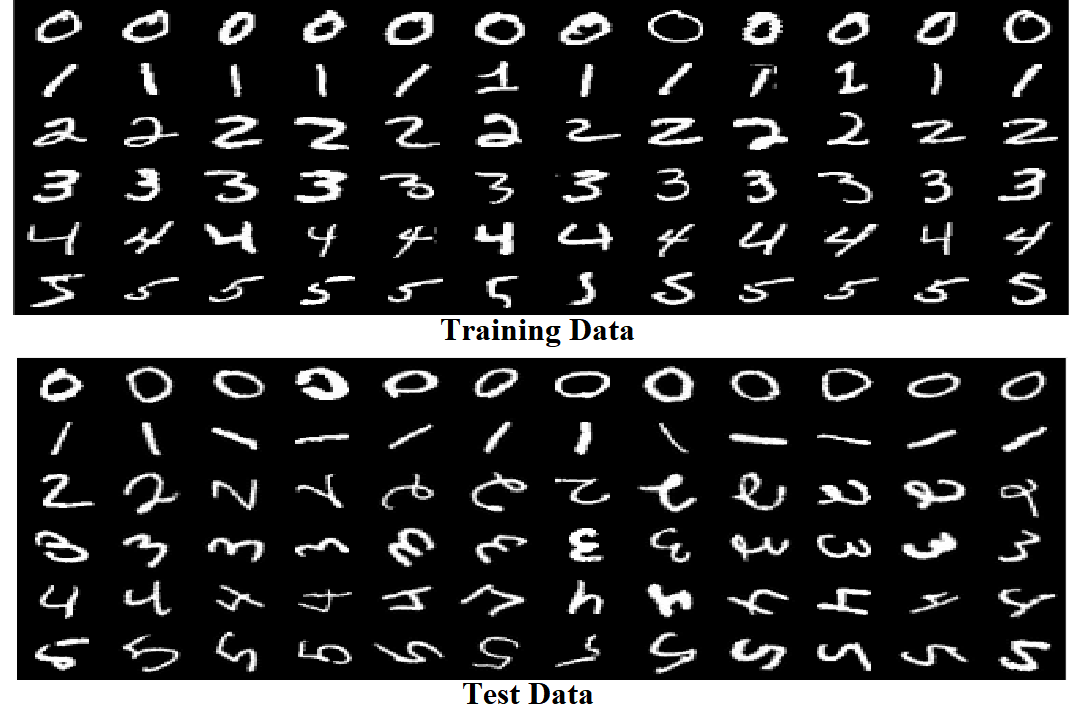}\label{figure:5(a)}\hfill}~~~~
	\subfloat[The OuTex\_00012 dataset.]
	{\includegraphics[height=45mm,width=65mm]{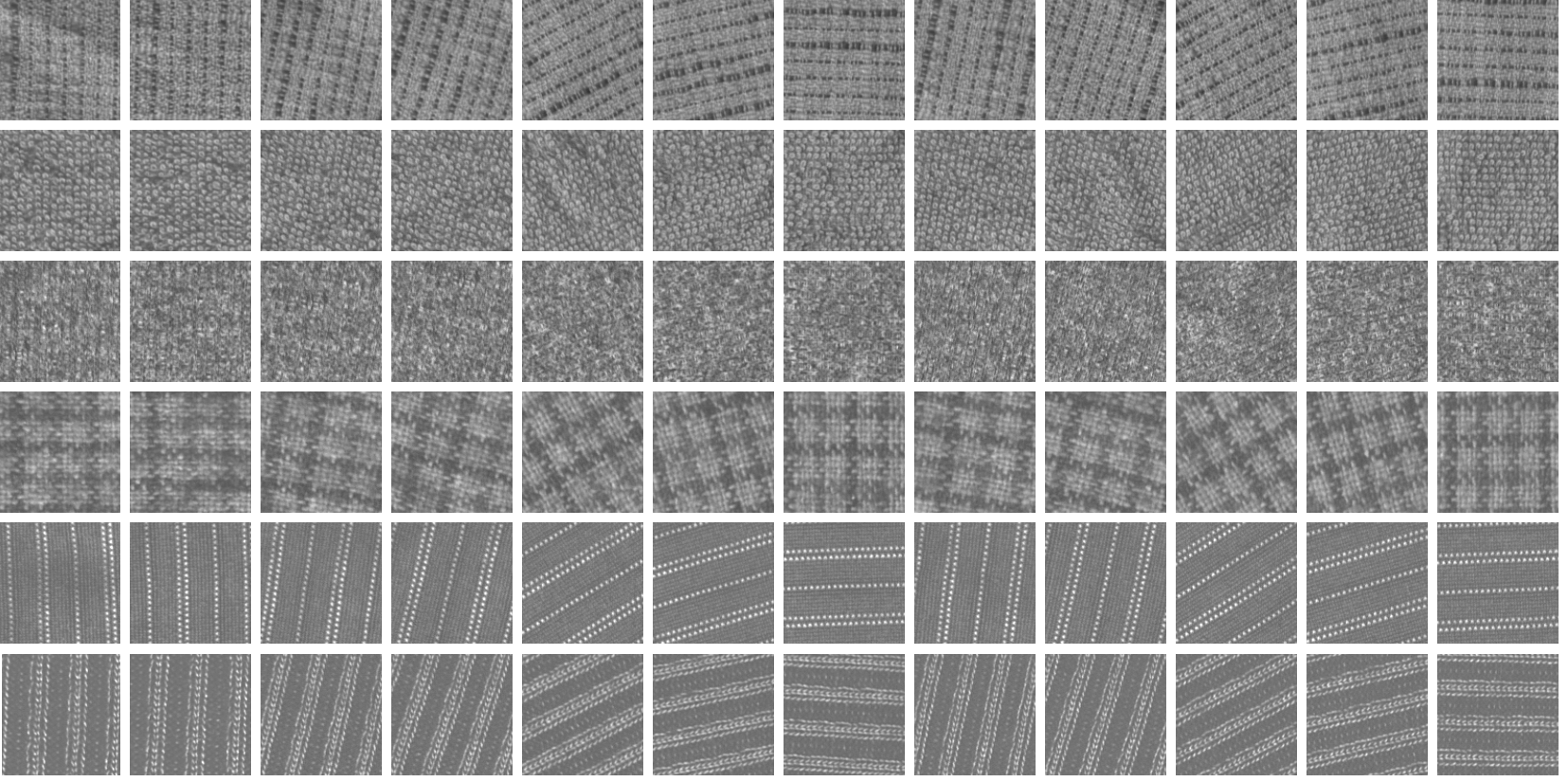}\label{figure:5(b)}\hfill}\\
	\subfloat[The MTARSI dataset.]
	{\includegraphics[height=35mm,width=65mm]{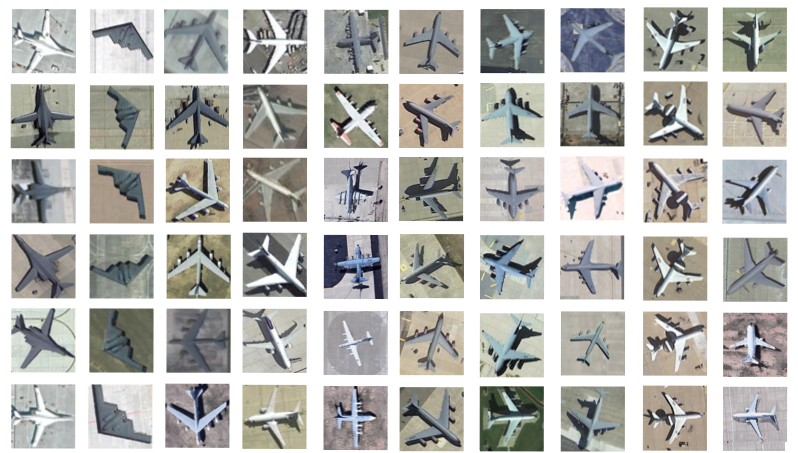}\label{figure:5(c)}\hfill}~~~~
	\subfloat[The NWPU-RESISC-45 dataset.]
	{\includegraphics[height=35mm,width=65mm]{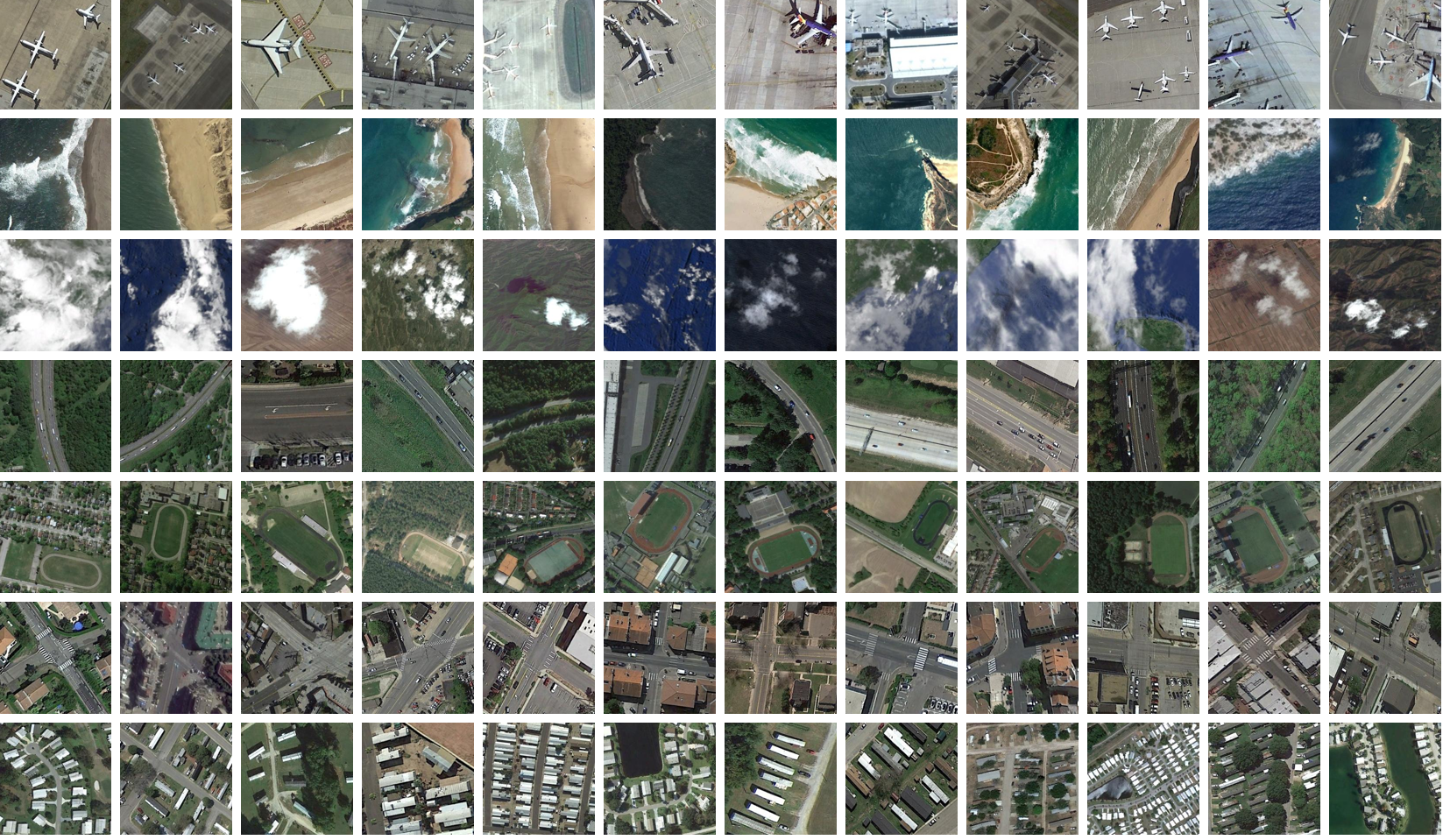}\label{figure:5(d)}\hfill}
	\vspace{-0.3cm}
	\caption{Various datasets used for evaluating the performance of RIConvs.}\label{figure:5}
\end{figure*}

\textbf{Datasets:} \textbf{(1)} \textbf{MNIST dataset} contains 70K $28\times28$ images of handwritten digits (0-9), with 60K designated for training and 10K for testing \cite{31}. Each test image is rotated 36 times, with rotation angles at $0^{\circ}, 10^{\circ}, 20^{\circ},...$, up to $360^{\circ}$, thus generating a total of 360K rotated test images. These form a new test set called MNIST-rot. Additionally, 10K training images are randomly selected for validation, leaving 50K images in the training set. Part of the images from the MNIST training set and the MNIST-Rot test set are shown in Fig.\ref{figure:5(a)}. \textbf{(2)} The Outex\_TC\_00012 dataset\cite{34} contains $9120$ texture images belonging to 24 different categories (all are $128\times128$ grayscale images). For each category, the collectors select $20$ texture surfaces of that category and initially photographed them under three lighting conditions ("inca", "t184", and "horizon"), which serve as training images. Then, under "t184" and "horizon" lighting conditions, the collectors photograph each texture surface again from 8 different angles ($5^{\circ}\sim90^{\circ}$) to serve as test images. Therefore, the training set includes $1440$ images, and the test set comprises $7680$ images. Part of the images from the training and test sets are shown in Fig.\ref{figure:2(b)}. \textbf{(3)} The MTARSI dataset is an RGB image dataset for aircraft type recognition tasks \cite{35}, containing $9385$ images of aircraft belonging to $20$ different types. As shown in Fig.\ref{figure:5(c)}, aircrafts in these images often have different orientations. Also, there are significant variances in the background. The size of each image was adjusted to $128\times128$, and then $200$ images are randomly selected from each category as training samples. Thus, the training set totals $4000$ images, with the remaining $5385$ images forming the test set. \textbf{(4)} NWPU-RESISC45 is a dataset for remote sensing image scene classification \cite{36}. It contains $31500$ RGB images belonging to $45$ scene categories, with each category comprising $700$ samples. All images were resized to $128\times128$, and $400$ images from each category are randomly selected as training images, with the remaining images used for testing. As shown in Fig.\ref{figure:5(d)}, due to arbitrary shooting angles, many categories such as “bridges” and “ground track fields” exhibit rotation variations. 

\textbf{Model:} \textbf{(1)} First, we need to verify the rotation invariance of all RConvs based on the MNIST dataset. For this purpose, we initially design a CNN model with six convolutional layers as the Baseline, with the number of kernels in each layer being $32$, $32$, $64$, $64$, $128$, and $128$, respectively. The kernel size of the last two convolutional layers is $3\times3$, while that of the first four layers is $7\times7$. After the second and fourth layers, there is a $2\times2$ max pooling operation, and after the last convolutional layer, there is a $7\times7$ average pooling layer. Since the size of the input images is $28\times28$, these pooling operations reduce the spatial size of the feature images to $1\times1$, reducing them to feature vectors. This vector is then fed into a fully connected layer with ten units for the final classification. By replacing all conventional convolution operations in this Baseline with certain type of RIConvs, we can obtain an RI-CNN model, such as SB-CNN, GD-CNN, and Max-CNN. \textbf{(2)} To illustrate that RIConvs can be integrated with common CNN backbones, we chose VGG16, Inception V1, DenseNet40, and ResNet18/34/50/101 as baselines. Similarly, by replacing all Convs in these baselines with RIConvs, we can obtain new RI-CNN models like GD-VGG16, Max-ResNet18, ST-DenseNet40, etc.

\textbf{Training protocol:} Our experiments are conducted on a Tesla V100 GPU (16G) with the Rocky Linux 8.7 system and PyTorch 2.0.0 framework. All models are trained from scratch without the use of pretrained parameters or data augmentation. This allows us to directly observe the performance improvement brought by RIConvs. \textbf{(1)} For experiments on the MNIST-Rot dataset, we train all baselines and RI-CNNs using the Adam optimizer, with an initial learning rate of $10^{-4}$, multiplied by 0.8 every 10 epochs. The number of epochs and batch size are both set to 100. \textbf{(2)} For experiments with the Outex\_00012, MTARSI, and NWPU-RESISC45 datasets, we also use the Adam optimizer, setting the number of epochs to 100 and batch size to 10. For VGG16, Inception V1, ResNet18/34/50/101, and the RI-CNNs derived from these backbones, the initial learning rate is set to $10^{-4}$. For DenseNet40 and its rotation-invariant models, the learning rate is set to $10^{-2}$. The value of the learning rate is reduced by a factor of $0.6$ every 10 epochs.

\subsection{Evaluating Rotation Invariance of RIConvs on the MNIST-Rot Dataset}
\label{Section4.2}
\begin{figure}
	\centering
	\includegraphics[width=140mm,height=75mm]{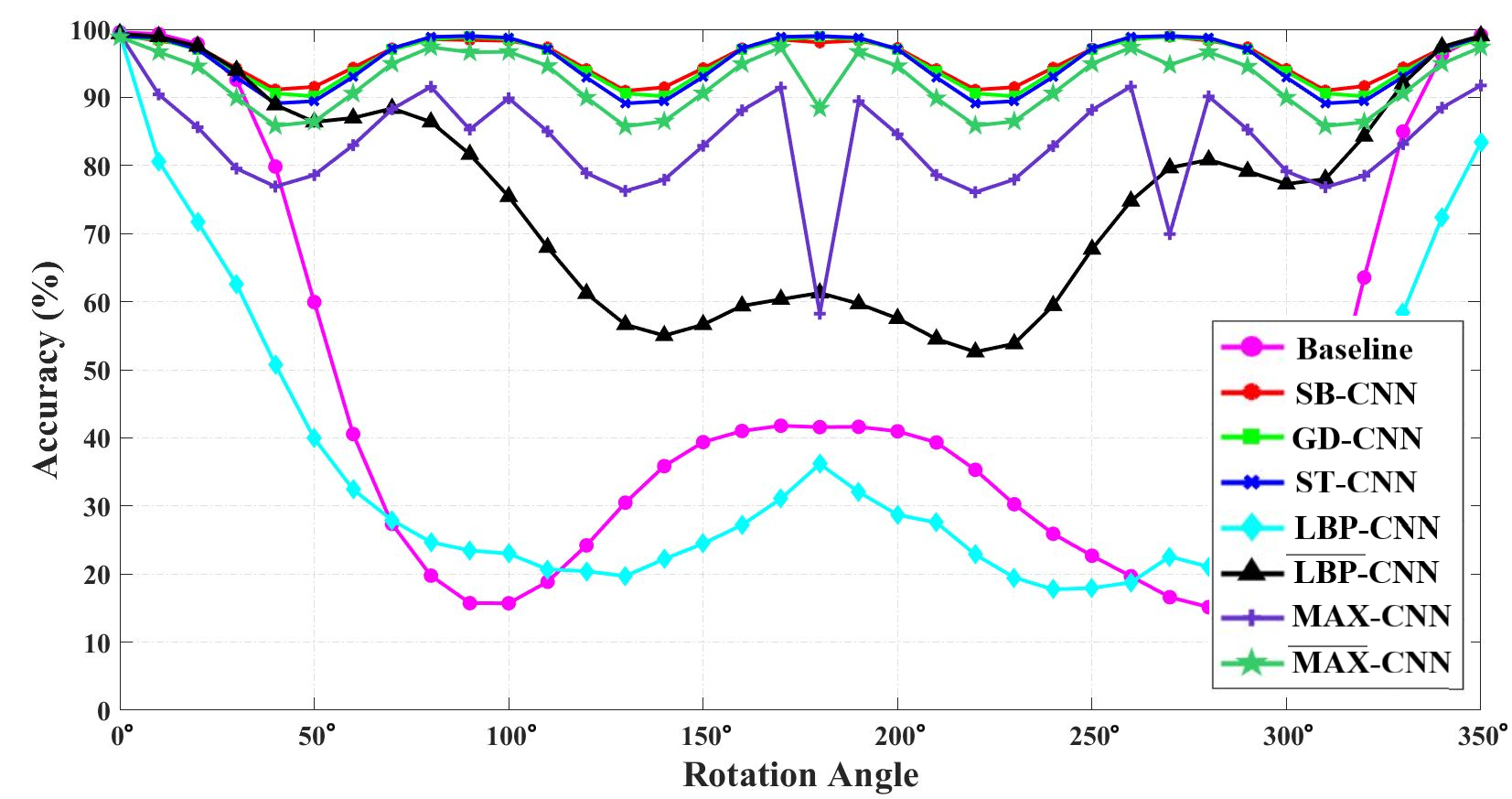}\\
	\caption{The classification accuracies from seven RIConvs on 36 rotated test subsets of MNIST-Rot with specific rotation angles ($0$, $10^{\circ}$, $20^{\circ}$,...,$350^{\circ}$).}\label{figure:6}
\end{figure}

Based on the MNIST dataset, we first validate rotation invariance of seven RIConvs. Without using data augmentation, we train seven types of RI-CNNs and their corresponding CNN baseline on the original MNIST training set. Then, we test the classification accuracy of these models on the MNIST-Rot test set. In fact, the MNIST-Rot test set contains $36$ subsets, each with $10000$ samples having the same rotation angle, such as $10^{\circ}$, $20^{\circ}$, etc. By observing the performance of the models on these subsets, we can analyze their invariance under different rotation angles. The final experimental results are shown in Fig.\ref{figure:6}. Firstly, we can find that SB-CNN, GD-CNN, ST-CNN, and $\overline{\mathrm{MAX}}$-CNN achieve more than $90\%$ accuracy at almost any rotation angle, of course, with $\overline{\mathrm{MAX}}$-CNN being slightly inferior to the first three. It is evident that their classification curves have a period of $90^{\circ}$. Taking the first period of $0^{\circ}\sim90^{\circ}$ as an example, the model's classification accuracy gradually decreases within the $0^{\circ}\sim45^{\circ}$ interval and gradually increases from $45^{\circ}\sim90^{\circ}$, reaching the performance minimum at a rotation angle of $45^{\circ}$. In fact, when rotating images, interpolation operations are needed; as the rotation angle increases from $0^{\circ}$ to $45^{\circ}$, the interpolation error gradually increases, and as the angle continues to increase from $45^{\circ}$ to $90^{\circ}$ the error gradually decreases. At a rotation angle of $90^{\circ}$, the interpolation operation does not introduce any computational error.

\vspace{-0.1cm}

Although the accuracy curve of MAX-CNN also shows a certain degree of periodicity, its performance is significantly worse than that of $\overline{\mathrm{MAX}}$-CNN. This indicates that, in most cases, the max operation cannot directly use pixel values to accurately estimate the orientation of a small region involved in convolution computations, whereas using the difference between adjacent pixels can achieve this. As we mentioned previous, the prerequisite for accurately estimating the orientation of a local region with the maximum operation is the existence of a unique maximum value within the region. However, this condition is not always met. When multiple positions in the region have the same pixel value, which is also the maximum value, we cannot use the maximum operation to achieve rotation invariance of the convolution. Calculating the difference between adjacent pixels can alleviate this problem to some extent. Suppose the pixel values at eight positions (excluding the center point) sampled in a $3\times3$ region based on the polar coordinate system are $0,0,0,0,1,1,1,1$, obviously, there are four identical maximum values in the region. But if we calculate the pixel difference between adjacent pixels, the result becomes $0,0,0,-1,0,0,0,1$. Clearly, there is only one maximum value $1$ at this time. It is for this reason that the performance of $\overline{\mathrm{MAX}}$-CNN is significantly better than that of MAX-CNN. 

The poor performance of $\overline{\mathrm{LBP}}$-CNN and LBP-CNN is caused by similar issues. For instance, when the binary pattern calculated within a $3\times3$ region exhibits periodicity, such as $01010101$, $00110011$, etc., the LBP operator cannot accurately determine the region's rotation angle. Replacing pixel difference between other positions and the central positions with the difference between adjacent positions can mitigate this issue, but only to a limited extent. Therefore, although $\overline{\mathrm{LBP}}$-CNN performs better than LBP-CNN, it is still significantly inferior to other methods. 

\begin{table}
	\caption{\label{table:1} The classification on original MNIST and MNIST-rot test sets. Bold stands for best results.}
	\centering
		\begin{tabular}{p{3.5cm}p{3cm}p{3cm}p{2.5cm}}
			\toprule[1.1pt]
			\textbf{Methods} & \textbf{Input Size} & \textbf{MNIST} & \textbf{MNIST-rot} \\
		    \toprule[1.1pt]
			ORN\cite{17} & $32\times32$ & \underline{$99.42\%$} & $80.01\%$\\
			RotEqNet\cite{19} & $28\times28$ & $99.26\%$ & $73.20\%$ \\
			G-CNN\cite{16} & $28\times 28$ & $99.27\%$ & $44.81\%$ \\
			H-Net\cite{12} & $32\times32$ & $99.19\%$ & $92.44\%$ \\
			B-CNN\cite{20} & $32\times32$ & $97.40\%$ & $88.29\%$ \\
			E(2)-CNN\cite{18} & $29\times 29$ & $98.14\%$ & $94.37\%$ \\
			\toprule[1.1pt]
			Baseline & $28\times 28$ & \bm{$99.57\%$} & $45.35\%$\\
			SB-CNN & $28\times 28$ & $99.08\%$ & \bm{$95.68\%$}\\
			GD-CNN & $28\times 28$ & $98.88\%$ & \underline{$95.35\%$}\\
			ST-CNN & $28\times 28$ & $99.02\%$ & $95.05\%$\\
			LBP-CNN & $28\times 28$ & $99.34\%$ & $36.31\%$\\
			$\overline{\mathrm{LBP}}$-CNN & $28\times 28$ & $99.36\%$ & $75.28\%$\\
			MAX-CNN & $28\times 28$ & $99.31\%$ & $83.32\%$\\
			$\overline{\mathrm{MAX}}$-CNN & $28\times 28$ & $98.85\%$ & $92.33\%$\\
			\midrule[1.1pt]
		\end{tabular}
\end{table} 

Table \ref{table:1} shows the classification accuracies of seven RIConvs, their CNN baseline, and six previously proposed RI-CNN models on the original MNIST test set and MNIST-rot test set. Without relying on data augmentation, Harmonic Network (H-Net)\cite{12}, Bessel CNN (B-CNN)\cite{20}, and E(2)-CNN \cite{18} also exhibit invariance to arbitrary rotation angles, while ORN\cite{17}, RotEqNet\cite{19}, and G-CNN\cite{16} only exhibit invariance to specific rotation angles such as $45^{\circ}$ or multiples of $90^{\circ}$. We directly train them using the code and protocols provided by their authors. Additionally, we do not compare with methods like STN \cite{14}, TI-Pooling \cite{26}, as their invariance relies on data augmentation. Our results indicate that: \textbf{(1)} on MNIST-rot, SB-CNN, GD-CNN, and ST-CNN surpass the previous state-of-the-art method E(2)-CNN, increasing the accuracy from $94.37\%$ to $95.68\%$, $95.35\%$, and $95.05\%$, respectively. MAX-CNN achieves an accuracy of $92.33\%$, comparable to H-Net. The performances of MAX-CNN and $\overline{\mathrm{LBP}}$-CNN are similar to those of ORB and RotEqNet, which exhibit invariance only to specific rotation angles, but significantly better than our baseline ($45.35\%$). The accuracy of LBP-CNN is only $36.31\%$, even worse than the baseline model. We have previously analyze the reasons. When the computed LBP based on regional information has periodicity, we cannot accurately estimate the orientation of that region. This situation occurs more frequently for binary images. \textbf{(2)} On the original MNIST test set, the baseline model achieves the best result ($99.57\%$). Previous research \cite{27,28,29} has shown that RI-CNNs have difficulty distinguishing certain digits, such as "9" and "6", which leads to slightly lower performance on this test set. This problem also results in stronger rotation-invariant models (such as SB-CNN, GD-CNN, and ST-CNN) performing worse than models with weaker rotation invariance (such as ORN, RotEqNet, and MAX-CNN) on the original test set.


\begin{table}
	\caption{\label{table:2} The classification on original MNIST and MNIST-rot test sets. Bold stands for best results.}
	\centering
	\setlength{\tabcolsep}{7pt}
	{\scriptsize
		\begin{tabular}{*{9}{c}}  
			\toprule[1.1pt]
			\multicolumn{9}{c}{\textbf{Training Data: 1440}} \\  
			\midrule  
			Model & \raisebox{-0.3ex}{/} & \raisebox{-0.5ex}{SB} & \raisebox{-0.5ex}{GD} & \raisebox{-0.5ex}{ST} & \raisebox{-0.5ex}{LBP} & $\raisebox{-0.5ex}{$\overline{\mathrm{LBP}}$}$ & \raisebox{-0.5ex}{MAX} & $\raisebox{-0.5ex}{$\overline{\mathrm{MAX}}$}$ \\ 
			\midrule  
			VGG & 62.40\% & \underline{89.56\%} & 87.93\% & \textbf{95.43\%} & 77.24\% & 81.25\% & 83.66\% & 84.39\% \\ 
			Inception V1 & 66.30\% & \underline{92.23\%} & 90.46\% & \textbf{94.39\%} & 80.63\% & 72.11\% & 82.68\% & 81.61\% \\
			DenseNet40 & 65.91\% & 89.32\% & \underline{92.24\%} & \textbf{94.23\%} & 88.65\% & 89.92\% & 78.89\% & 88.67\% \\
			ResNet18 & 63.18\% & \underline{89.19\%} & 88.98\% & \textbf{95.63\%} & 78.31\% & 77.89\% & 83.33\% & 80.53\% \\
			ResNet34 & 63.50\% & 91.05\% & \underline{91.33\%} & \textbf{95.35\%} & 79.73\% & 75.82\% & 86.86\% & 82.89\% \\
			ResNet50 & 69.80\% & \underline{92.50\%} & 91.00\% & \textbf{96.12\%} & 83.66\% & 79.39\% & 90.59\% & 85.53\% \\
			ResNet101 & 70.96\% & \underline{93.57\%} & 89.83\% & \textbf{95.33\%} & 82.38\% & 82.98\% & 93.52\% & 88.09\% \\
			\midrule
			\multicolumn{9}{c}{\textbf{Training Data: 960}} \\  
			\midrule  
			Model & \raisebox{-0.3ex}{/} & \raisebox{-0.5ex}{SB} & \raisebox{-0.5ex}{GD} & \raisebox{-0.5ex}{ST} & \raisebox{-0.5ex}{LBP} & $\raisebox{-0.5ex}{$\overline{\mathrm{LBP}}$}$ & \raisebox{-0.5ex}{MAX} & $\raisebox{-0.5ex}{$\overline{\mathrm{MAX}}$}$ \\ 
			\midrule  
			VGG & 61.84\% & \underline{90.10\%} & 88.20\% & \textbf{95.95\%} & 79.99\% & 78.27\% & 83.82\% & 80.47\% \\ 
			Inception V1 & 64.73\% & \underline{90.16\%} & 89.19\% & \textbf{95.79\%} & 78.29\% & 74.51\% & 82.20\% & 84.15\% \\
			DenseNet40 & 70.13\% & \underline{93.22\%} & 85.14\% & \textbf{96.09\%} & 77.80\% & 76.77\% & 79.97\% & 83.05\% \\
			ResNet18 & 64.44\% & \underline{91.61\%} & 90.66\% & \textbf{96.47\%} & 82.94\% & 70.30\% & 84.65\% & 82.37\% \\
			ResNet34 & 63.31\% & \underline{94.75\%} & 89.34\% & \textbf{96.21\%} & 86.28\% & 77.50\% & 85.87\% & 81.22\% \\
			ResNet50 & 70.31\% & 90.79\% & \underline{92.84\%} & \textbf{97.30\%} & 84.14\% & 83.02\% & 88.55\% & 87.88\% \\
			ResNet101 & 67.42\% & \underline{92.17\%} & 89.40\% & \textbf{94.23\%} & 85.63\% & 82.92\% & 86.33\% & 85.01\% \\
			\midrule
			\multicolumn{9}{c}{\textbf{Training Data: 480}} \\  
			\midrule  
			Model & \raisebox{-0.3ex}{/} & \raisebox{-0.5ex}{SB} & \raisebox{-0.5ex}{GD} & \raisebox{-0.5ex}{ST} & \raisebox{-0.5ex}{LBP} & $\raisebox{-0.5ex}{$\overline{\mathrm{LBP}}$}$ & \raisebox{-0.5ex}{MAX} & $\raisebox{-0.5ex}{$\overline{\mathrm{MAX}}$}$ \\ 
			\midrule  
			VGG & 60.09\% & 84.97\% & \underline{86.59\%} & \textbf{93.96\%} & 79.95\% & 71.12\% & 80.34\% & 81.50\% \\ 
			Inception V1 & 67.88\% & 86.84\% & \underline{88.23\%} & \textbf{93.52\%} & 79.97\% & 68.66\% & 81.47\% & 81.61\% \\
			DenseNet40 & 59.84\% & \underline{88.79\%} & 83.31\% & \textbf{93.03\%} & 77.36\% & 78.75\% & 81.41\% & 78.20\% \\
			ResNet18 & 63.57\% & 84.87\% & \underline{86.50\%} & \textbf{93.31\%} & 77.19\% & 68.84\% & 81.26\% & 78.80\% \\
			ResNet34 & 64.43\% & 84.54\% & \underline{86.35\%} & \textbf{93.31\%} & 78.67\% & 67.33\% & 87.76\% & 82.33\% \\
			ResNet50 & 66.02\% & \underline{86.12\%} & 84.84\% & \textbf{91.59\%} & 85.70\% & 82.12\% & 86.45\% & 84.15\% \\
			ResNet101 & 63.09\% & \underline{87.21\%} & 85.72\% & \textbf{90.25\%} & 80.14\% & 79.38\% & 86.24\% & 80.36\% \\
			\midrule[1.1pt]
		\end{tabular}
	}
\end{table}

\subsection{The performance of RIConvs in Real-World Image Classification} 
By replacing traditional convolutions with different types of RIConvs in seven common CNN backbones, we obtain a new set of rotation-invariant backbones. This section validates the performance of these rotation-invariant models on three real datasets: Outex\_TC\_00012, MTARSI, and NWPU-RESISC45. Theoretically, rotation invariance will enhance the feature extraction capability of the model, enabling it to learn key features of the object based on fewer training samples. To illustrate this point, we gradually reduce the number of samples in the training set while keeping the test set unchanged. Taking MTARSI as an example, we reduced the number of samples in the training set from $4000$ ($200$ per class) to $3000$ ($150$ per class) and $2000$ ($100$ per class). Fig.\ref{figure:4} shows the classification accuracies achieved by original backbones and corresponding RI-CNN models on the same test set when they are trained on different sizes of training sets. Analyzing the results obtained on the NWPU-RESISC dataset (see Fig.\ref{figure:4(c)}), we can observe the followings: \textbf{(1)} In most cases, RIConvs significantly improve the performance of various CNN backbones. For example, when trained with a the dataset containing $18000$ samples ($400$ per class), the original ResNet18 achieves a classification accuracy of $82.85\%$ on the test set, while all RI-ResNet18 models perform better. Specifically, SB-ResNet18, GD-ResNet, and $\overline{MAX}$-ResNet18 achieve classification accuracies of $90.63\%$, $90.92\%$, and $90.55\%$, respectively. Even the previously poorly performing LBP-Convs on the MNIST-Rot dataset also achieve an accuracy of $87.02\%$. \textbf{(2)} The fewer training samples, the more significant the performance improvement brought by RIConvs. For example, when the number of training samples is $18000$, compared to the original ResNet34 with a classification accuracy of $82.82\%$, using seven RI-ResNet34 (SB, GD, ST, LBP, $\overline{LBP}$, MAX, and $\overline{LBP}$) respectively achieve accuracy improvements of $7.39\%$, $7.35\%$, $7.19\%$, $3.28\%$, $5.81\%$, $7.06\%$, and $7.13\%$. When the training samples are reduced to $13000$, although the performance of all models decreases, the accuracy improvement brought by using seven RIConvs increases to $10.34\%$, $11.03\%$, $10.14\%$, $6.46\%$, $8.48\%$, $10.10\%$, and $10.54\%$. \textbf{(3)} RI-CNN models implemented based on SB-Conv, GD-Conv, and ST-Conv achieve better accuracy in most cases. This indicates that RIConvs implemented using gradient operators and sorting operations exhibit better invariance. Additionally, compared to the experimental results on the MNIST-Rot dataset, all RIConvs perform relatively well on the NWPU-RESISC45 dataset, including MAX-Conv and LBP-Conv. This is because local patterns in binary images tend to be simpler, making it easier to encounter issues such as multiple identical maximum values and periodicity in computed LBP features, while local patterns in color image data are more complex, reducing the likelihood of encountering such issues. The above observations can also be made based on the experimental results obtained on the other two datasets (see Fig.\ref{figure:4(a)} and Fig.\ref{figure:4(b)}).  

\begin{table}
	\caption{\label{table:3} The classification on original MNIST and MNIST-rot test sets. Bold stands for best results.}
	\centering
	\setlength{\tabcolsep}{5pt}
	{\scriptsize
		\begin{tabular}{*{9}{c}}  
			\toprule[1.1pt]
			\multicolumn{9}{c}{\textbf{Training Data: 4000}} \\  
			\midrule  
			Model & \raisebox{-0.3ex}{/} & \raisebox{-0.5ex}{SB} & \raisebox{-0.5ex}{GD} & \raisebox{-0.5ex}{ST} & \raisebox{-0.5ex}{LBP} & $\raisebox{-0.5ex}{$\overline{\mathrm{LBP}}$}$ & \raisebox{-0.5ex}{MAX} & $\raisebox{-0.5ex}{$\overline{\mathrm{MAX}}$}$ \\ 
			\midrule  
			VGG & 95.17\% & 96.51\% & \textbf{97.29\%} & 96.86\% & 94.44\% & 96.58\% & \underline{96.90\%} & 96.75\% \\ 
			Inception V1 & 96.49\% & \underline{96.67\%} & 97.66\% & 97.57\% & 95.67\% & 97.14\% & 97.57\% & \textbf{97.68\%} \\
			DenseNet40 & 93.22\% & 94.07\% & 93.77\% & \underline{95.35\%} & 94.20\% & 94.28\% & \textbf{95.46\%} & 93.36\% \\
			ResNet18 & 87.42\% & 96.34\% & \textbf{97.42\%} & \underline{96.99\%} & 94.07\% & 96.25\% & 96.77\% & 96.67\% \\
			ResNet34 & 89.55\% & 96.91\% & \textbf{97.32\%} & 96.56\% & 94.31\% & 96.21\% & \underline{96.93\%} & 96.71\% \\
			ResNet50 & 90.22\% & 96.02\% & \textbf{96.93\%} & 95.89\% & 93.35\% & 95.46\% & 96.19\% & \underline{96.56\%} \\
			ResNet101 & 91.06\% & 95.33\% & \textbf{96.62\%} & 95.30\% & 93.62\% & 95.82\% & \underline{95.93\%} & 95.72\% \\
			\midrule
			\multicolumn{9}{c}{\textbf{Training Data: 3000}} \\  
			\midrule  
			Model & \raisebox{-0.3ex}{/} & \raisebox{-0.5ex}{SB} & \raisebox{-0.5ex}{GD} & \raisebox{-0.5ex}{ST} & \raisebox{-0.5ex}{LBP} & $\raisebox{-0.5ex}{$\overline{\mathrm{LBP}}$}$ & \raisebox{-0.5ex}{MAX} & $\raisebox{-0.5ex}{$\overline{\mathrm{MAX}}$}$ \\ 
			\midrule  
			VGG & 92.27\% & 93.75\% & 95.17\% & 95.04\% & 91.93\% & 94.35\% & \underline{95.27\%} & \textbf{95.59\%} \\ 
			Inception V1 & 93.42\% & 94.85\% & \underline{95.87\%} & \textbf{96.30\%} & 93.66\% & 94.89\% & 95.69\% & 95.84\% \\
			DenseNet40 & 89.20\% & 90.26\% & 92.36\% & \textbf{93.51\%} & 86.19\% & 92.43\% & 92.97\% & \underline{93.20\%} \\
			ResNet18 & 83.14\% & 94.11\% & \textbf{95.30\%} & \textbf{95.30\%} & 90.33\% & 93.68\% & 94.01\% & \underline{94.39\%} \\
			ResNet34 & 85.84\% & 94.33\% & \textbf{95.65\%} & 94.68\% & 91.93\% & 94.57\% & 95.06\% & \underline{95.46\%} \\
			ResNet50 & 85.95\% & 92.84\% & \textbf{94.93\%} & 93.75\% & 89.89\% & 92.55\% & 94.00\% & \underline{94.03\%} \\
			ResNet101 & 88.98\% & 93.57\% & \textbf{94.16\%} & 91.90\% & 90.45\% & 92.90\% & \underline{93.48\%} & 93.10\% \\
			\midrule
			\multicolumn{9}{c}{\textbf{Training Data: 2000}} \\  
			\midrule  
			Model & \raisebox{-0.3ex}{/} & \raisebox{-0.5ex}{SB} & \raisebox{-0.5ex}{GD} & \raisebox{-0.5ex}{ST} & \raisebox{-0.5ex}{LBP} & $\raisebox{-0.5ex}{$\overline{\mathrm{LBP}}$}$ & \raisebox{-0.5ex}{MAX} & $\raisebox{-0.5ex}{$\overline{\mathrm{MAX}}$}$ \\ 
			\midrule  
			VGG & 88.23\% & 90.11\% & \textbf{92.51\%} & \underline{91.91\%} & 86.51\% & 89.31\% & 91.25\% & 91.58\% \\ 
			Inception V1 & 88.42\% & 91.51\% & 93.48\% & \underline{93.90\%} & 89.18\% & 92.30\% & \textbf{94.15\%} & 93.35\% \\
			DenseNet40 & 83.68\% & 86.17\% & 87.57\% & \underline{88.62\%} & 80.39\% & 88.18\% & 87.29\% & \textbf{89.35\%} \\
			ResNet18 & 76.34\% & 90.15\% & \underline{92.06\%} & \textbf{92.73\%} & 85.13\% & 90.00\% & 90.91\% & 90.61\% \\
			ResNet34 & 79.87\% & 90.86\% & \underline{92.19\%} & 91.62\% & 86.69\% & 90.74\% & \textbf{92.40\%} & 91.91\% \\
			ResNet50 & 80.22\% & 89.03\% & \textbf{91.97\%} & 90.30\% & 83.29\% & 89.57\% & 90.13\% & \underline{90.82\%} \\
			ResNet101 & 81.80\% & 86.78\% & \textbf{90.00\%} & 87.57\% & 83.03\% & 88.42\% & \underline{88.72\%} & 88.03\% \\
			\midrule[1.1pt]
		\end{tabular}
	}
\end{table}

\begin{table}
	\caption{\label{table:4} The classification on original MNIST and MNIST-rot test sets. Bold stands for best results.}
	\centering
	\setlength{\tabcolsep}{5pt}
	{\scriptsize
		\begin{tabular}{*{9}{c}}  
			\toprule[1.1pt]
			\multicolumn{9}{c}{\textbf{Training Data: 18000}} \\  
			\midrule  
			Model & \raisebox{-0.3ex}{/} & \raisebox{-0.5ex}{SB} & \raisebox{-0.5ex}{GD} & \raisebox{-0.5ex}{ST} & \raisebox{-0.5ex}{LBP} & $\raisebox{-0.5ex}{$\overline{\mathrm{LBP}}$}$ & \raisebox{-0.5ex}{MAX} & $\raisebox{-0.5ex}{$\overline{\mathrm{MAX}}$}$ \\ 
			\midrule  
			VGG & 86.16\% & 88.82\% & \underline{89.63\%} & 89.55\% & 85.81\% & 88.40\% & 89.39\% & \textbf{89.72\%} \\ 
			Inception V1 & 88.99\% & 91.09\% & 91.44\% & 91.77\% & 89.60\% & 90.64\% & \underline{91.54\%} & \textbf{91.68\%} \\
			DesNet40 & 84.14\% & \textbf{86.61\%} & 85.78\% & 86.24\% & \underline{86.52\%} & 85.64\% & 84.22\% & 86.30\% \\
			ResNet18 & 82.85\% & \underline{90.63\%} & \textbf{90.92\%} & 89.99\% & 87.02\% & 89.11\% & 89.93\% & 90.55\% \\
			ResNet34 & 82.82\% & \textbf{90.21\%} & \underline{90.17\%} & 90.01\% & 86.10\% & 88.63\% & 89.88\% & 89.95\% \\
			ResNet50 & 81.82\% & \underline{89.45\%} & \textbf{89.47\%} & 88.31\% & 84.29\% & 87.48\% & 88.14\% & 88.50\% \\
			ResNet101 & 82.93\% & \underline{86.15\%} & 85.56\% & 81.21\% & 83.11\% & 83.21\% & \textbf{87.05\%} & 84.11\% \\
			\midrule
			\multicolumn{9}{c}{\textbf{Training Data: 13500}} \\  
			\midrule  
			Model & \raisebox{-0.3ex}{/} & \raisebox{-0.5ex}{SB} & \raisebox{-0.5ex}{GD} & \raisebox{-0.5ex}{ST} & \raisebox{-0.5ex}{LBP} & $\raisebox{-0.5ex}{$\overline{\mathrm{LBP}}$}$ & \raisebox{-0.5ex}{MAX} & $\raisebox{-0.5ex}{$\overline{\mathrm{MAX}}$}$ \\ 
			\midrule  
			VGG & 83.89\% & 86.81\% & \textbf{87.58\%} & \underline{87.41\%} & 83.30\% & 85.11\% & 86.61\% & 86.92\% \\ 
			Inception V1 & 87.07\% & 89.44\% & 89.93\% & \underline{90.43\%} & 87.10\% & 89.53\% & \textbf{90.88\%} & 90.38\% \\ 
			DesNet40 & 83.10\% & 83.59\% & \textbf{84.88\%} & \underline{84.45\%} & 80.13\% & 83.80\% & 82.90\% & 84.08\% \\ 
			ResNet18 & 80.16\% & \textbf{88.96\%} & \textbf{88.96\%} & \underline{88.93\%} & 84.68\% & 87.13\% & 88.46\% & 88.65\% \\ 
			ResNet34 & 77.73\% & 88.07\% & \textbf{88.76\%} & 87.87\% & 84.19\% & 86.21\% & 87.83\% & \underline{88.27\%} \\ 
			ResNet50 & 78.96\% & \underline{87.13\%} & \textbf{87.47\%} & 84.39\% & 80.90\% & 84.27\% & 85.15\% & 85.75\% \\ 
			ResNet101 & 78.73\% & 84.81\% & \textbf{86.07\%} & 82.96\% & 80.19\% & 83.56\% & 85.17\% & \underline{85.26\%} \\ 
			\midrule
			\multicolumn{9}{c}{\textbf{Training Data: 9000}} \\  
			\midrule  
			Model & \raisebox{-0.3ex}{/} & \raisebox{-0.5ex}{SB} & \raisebox{-0.5ex}{GD} & \raisebox{-0.5ex}{ST} & \raisebox{-0.5ex}{LBP} & $\raisebox{-0.5ex}{$\overline{\mathrm{LBP}}$}$ & \raisebox{-0.5ex}{MAX} & $\raisebox{-0.5ex}{$\overline{\mathrm{MAX}}$}$ \\ 
			\midrule  
			VGG & 79.67\% & 82.90\% & \textbf{84.17\%} & 83.57\% & 79.73\% & 81.76\% & 83.01\% & \underline{84.09\%} \\ 
			Inception V1 & 83.76\% & 86.53\% & 86.73\% & \textbf{88.04\%} & 83.63\% & 85.36\% & 86.94\% & \underline{87.24\%} \\ 
			DesNet40 & 77.18\% & \textbf{82.22\%} & \underline{81.79\%} & 81.00\% & 76.16\% & 80.44\% & 81.03\% & 80.10\% \\ 
			ResNet18 & 72.02\% & \underline{85.66\%} & \textbf{86.41\%} & 85.44\% & 81.81\% & 83.59\% & 85.18\% & 85.61\% \\ 
			ResNet34 & 71.47\% & 84.76\% & \textbf{85.69\%} & 84.27\% & 80.92\% & 83.68\% & 84.48\% & \underline{84.96\%} \\ 
			ResNet50 & 70.65\% & \textbf{83.63\%} & \underline{83.56\%} & 80.26\% & 77.24\% & 81.38\% & 81.32\% & 82.23\% \\ 
			ResNet101 & 71.34\% & \underline{82.60\%} & \textbf{83.55\%} & 78.93\% & 76.06\% & 80.41\% & 81.16\% & 81.10\% \\ 
			\midrule[1.1pt]
		\end{tabular}
	}
\end{table}

\subsection{The Influence of Data Augmentation}
Finally, we analyze the impact of data augmentation on the performance of RIConvs. Using different sizes of the MTARSI training set, we obtain the classification accuracies of ResNet18 and seven RI-ResNet18 models on the test set. Unlike the experiment setting in previous sections, this time we apply data augmentation during model training, where each input image was randomly rotated by an angle within $(0, 2\pi]$. The results are shown in Table\ref{table:2}. Comparing Table\ref{table:2} with Fig.\ref{figure:4(b)}, we can observe a significant improvement in performance for both ResNet18 and all RI-ResNet18 when data augmentation is employed. For example, when the training set contained 3000 images ($150$ per class), without data augmentation, the classification accuracies of ResNet18 and GD-ResNet18 are $83.14\%$ and $95.30\%$, respectively. With data augmentation, their accuracies improve to $93.83\%$ and $96.97\%$, demonstrating a notable enhancement. More importantly, we find that even when the models are trained with data augmentation, most RI-ResNet18 models (except LBP-ResNet18) still outperform the original ResNet18, especially when the training data is limited. This once again underscores the significance of constructing rotation-invariant convolutional operations. In fact, mechanism-assured rotation invariance endows CNN models with stronger feature learning capabilities and higher utilization efficiency of learnable parameters. Therefore, even when data augmentation is applied during training, this property further enhances the model's performance.

\begin{table}
	\caption{\label{table:5} The performance of ResNets and seven RI-ResNet18 trained with data augmentation on the MTARSI test set. Bold stands for best results.}
	\centering
		\begin{tabular}{p{3.5cm}p{2.8cm}p{2.8cm}p{2.8cm}}
			\toprule[1.1pt]
			\textbf{Training Data} & \textbf{20}\bm{$\times$}\textbf{200=4K} & \textbf{20}\bm{$\times$}\textbf{150=3K} & \textbf{20}\bm{$\times$}\textbf{100=2K} \\
			\toprule[1.1pt]
			ResNet18 & $93.83\%$ & $90.89\%$ & $85.99\%$\\
			SB-ResNet18 & $95.11\%$ & $92.88\%$ & $88.36\%$ \\
			GD-ResNet18 & \bm{$96.97\%$} & \bm{$94.83\%$} & $90.06\%$ \\
			ST-ResNet18 & $96.02\%$ & $93.92\%$ & \bm{$90.32\%$} \\
			LBP-ResNet18 & $92.99\%$ & $88.40\%$ & $83.31\%$ \\
			$\overline{\mathrm{LBP}}$-ResNet18 & $95.17\%$ & $92.83\%$ & $88.96\%$ \\
			MAX-ResNet18 & $95.35\%$ & $93.25\%$ & $89.31\%$\\
			$\overline{\mathrm{MAX}}$-ResNet18 & $95.30\%$ & $93.44\%$ & $89.28\%$\\
			\midrule[1.1pt]
		\end{tabular}
\end{table} 

\section{Conclutions}
Based on non-learnable operations, this paper designs a set of convolutional operations that are invariant to arbitrary rotations. They have the same number of learnable parameters as their corresponding traditional convolutions and can be interchangeable. Using the MNIST-Rot dataset, we first verify the invariance of these RIConvs under different rotation angles, analyze the shortcomings of some non-learnable operations, and compare new RIConvs with existing RI-CNN models. Two types of RIConvs designed based on gradient operators achieve state-of-the-art results on this dataset. Subsequently, we combine these RIConvs with classic CNN backbones to obtain different RI-CNN models and validate their performance on texture classification, aircraft type recognition, and remote sensing image classification tasks based on the OuTex\_00012, MTARSI, and NWPU-RESISC45 datasets. Our experimental results demonstrate that the proposed RIConvs can effectively improve the performance of classic CNN models, especially when the training data is limited, regardless of whether data augmentation is used during training. In the future, we plan to further apply RIConvs to object detection and image segmentation tasks.





\end{document}